\documentclass[letterpaper, 10 pt, conference]{ieeeconf}  

\IEEEoverridecommandlockouts                              

\usepackage{graphicx}
\usepackage{amsmath} 
\usepackage{xcolor}
\usepackage{wrapfig}
\usepackage{amsfonts}
\usepackage{url}
\usepackage{paralist}

\title{\LARGE \bf
Learning to Prune Branches in Modern  Tree-Fruit Orchards

\author{Abhinav Jain$^{1}$, Cindy Grimm$^{1}$, Stefan Lee$^{1}$}
\thanks{$^{1}$Collaborative Robotics and Intelligent Systems (CoRIS) Institute, 
        Oregon State University, Corvallis OR 97331, USA
        {\tt\small \{jainab, grimmc, leestef\}@oregonstate.edu}}%
\thanks{This paper has been accepted for publication at the IEEE International Conference on Robotics and Automation (ICRA) 2025.}%
}

\newcommand{\perpproj}[2]{#1_{\perp #2}}

\begin{document}

\maketitle

\thispagestyle{empty}
\pagestyle{empty}

\begin{abstract}
Dormant tree pruning is labor-intensive but essential to maintaining modern highly-productive fruit orchards. In this work we present a closed-loop visuomotor controller for robotic pruning. The controller guides the cutter through a cluttered tree environment to reach a specified cut point and ensures the cutters are perpendicular to the branch. We train the controller using a novel orchard simulation that captures the geometric distribution of branches in a target apple orchard configuration. Unlike traditional methods requiring full 3D reconstruction, our controller uses just optical flow images from a wrist-mounted camera. We deploy our learned policy in simulation and the real-world for an example V-Trellis envy tree with zero-shot transfer, achieving a $\sim$30\% success rate --  approximately half the performance of an oracle planner.

\end{abstract}


\section{Introduction}
\vspace{-4pt}
\label{sec:introduction}

\begin{figure*}[h]
    \centering
    \vspace{0.5mm}
    \includegraphics[width = 0.9\textwidth, height = 4.2cm]{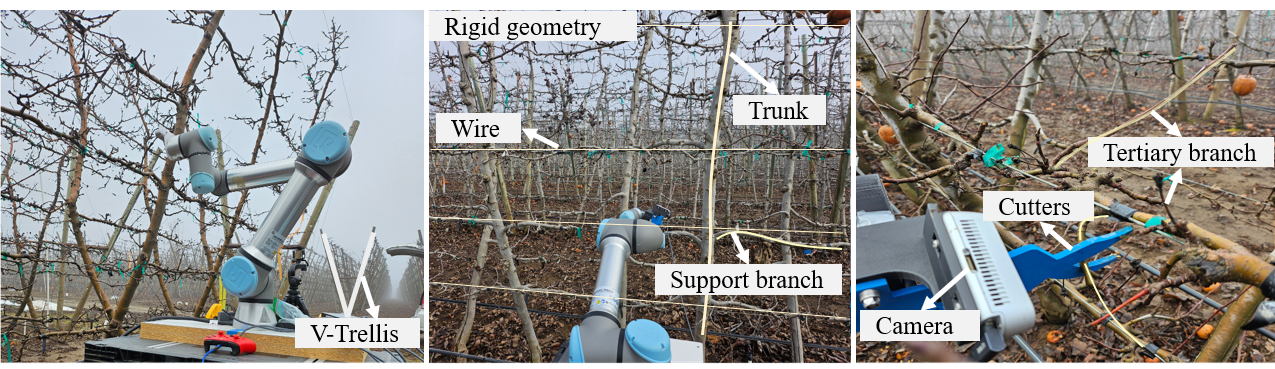}
    \vspace{-11pt}
    \caption{Pruning in modern, planar fruit orchards. Left and middle: V-Trellis architecture (Envy apples) from a modern orchard. Trees are grown 4' apart with branches tied down to wires spaced 18" apart (UR5 shown for scale). Right: Pruning setup, with an eye-in-hand camera mounted on the wrist (note: 3D printed version of the cutter used for safety in trials).}
    \label{fig:ur5_pruning_field}
    \vspace{-11pt}
\end{figure*}

Modern farming techniques have adopted carefully designed tree structures that improve productivity and labor efficiency but must be maintained through detailed dormant tree pruning and training. We focus on one such structure --- Envy apple trees in a V-trellis setting --- where trees are grown in approximately planar rows. The main trunk grows 15 degrees off vertical, and the primary support branches are tied to horizontal wires between posts (see Figure~\ref{fig:ur5_pruning_field}).

Unfortunately, dormant tree pruning is labor-intensive, representing up to 25\% of the annual labor costs in high-density apple orchards~\cite{pruningcost,wsuPruning,infaco2022farm}. These costs are projected to rise due to declining immigration of farm laborers to the United States, who have historically fulfilled this agricultural labor demand~\cite{nae2019immigrants,agamerica2022labor,daniels2018strawberries}.

Robotic pruning has the potential to address this labor shortage and produce consistent and reproducible pruning outcomes. However, there are numerous challenges to building a holistic system that makes this possible: i) perceiving the tree structure, ii) determining which branches to cut and where to cut them, and iii) driving the branch-cutting robot to the desired location to execute the cut {\em without} damaging rigid tree structures or supports. This paper's focus is on this last step -- learning a vision-based control policy to control a robotic pruner.\looseness=-1
\begin{figure}[t]
    \centering
    \includegraphics[width=0.48\textwidth]{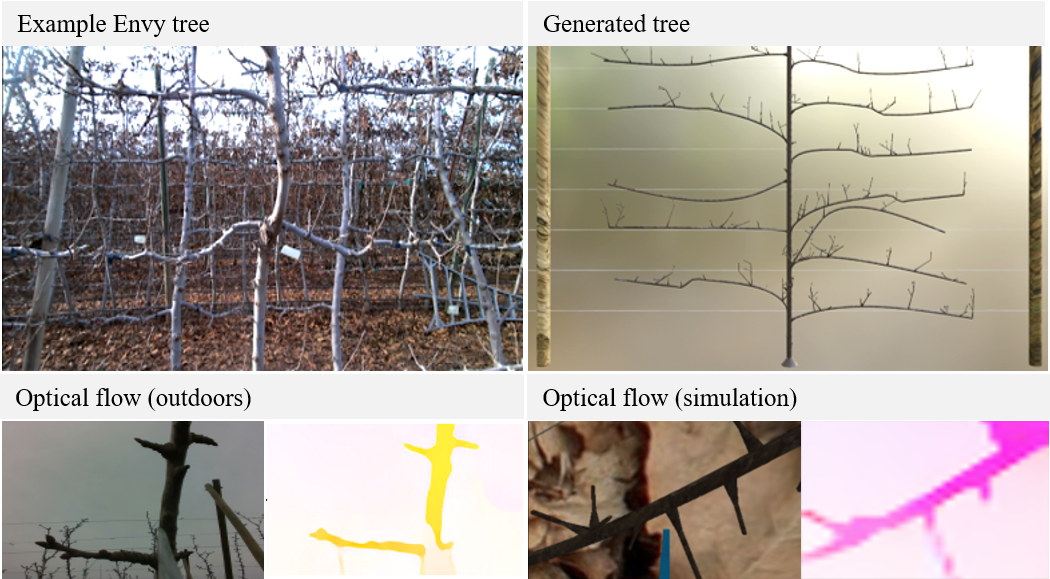}
    \caption{ Top-Left : An example Envy tree from an orchard. Top-Right: An example synthetic tree. Bottom-left: Image and corresponding optical flow generated outdoors. Bottom-Right: Example rendered image and corresponding optical flow.}
    \label{fig:tree_and_of}
    \vspace{-11pt}
\end{figure}

Robotic pruning in modern fruiting tree orchards (e.g., apple orchards) presents a challenging setting for robotic control. Individual tree geometries are unique and can be intricate -- introducing high variation and significant visual and physical clutter. Classical motion planning techniques use accurate 3D reconstruction of the tree to perform the pruning task\cite{CorbettDaviesStraddleVines,trimbot,SilwalBumblebee}. However, depth sensing technologies that rely on structured-light or coded-light methods struggle in outdoor environments due to interference from sunlight and surface scattering~\cite{lidaroutdoorbad}. Additionally, thin branches and a dynamic outdoor environment make this reconstruction problem difficult. Further, pruning cuts must be made at particular angles relative to the branch and require a wider range of target end-effector orientations and obstacle avoidance than standard reaching tasks. Despite this complexity, effective solutions must operate quickly to make an impact -- human pruners average one cut per second and roughly 10-50 cuts per tree~\cite{DeannaPruningDecisions}. As such, computationally intensive 3D reconstruction pipelines or planning techniques may result in solutions that are inaccurate and cost-prohibitive.

In this work we present a learning-based visuomotor control policy that guides a 6-DOF robotic arm with a cutter attachment to reach a given pruning point. The policy uses images from a wrist-mounted camera to avoid collisions and orient the cutter perpendicular to the branch, eliminating the need for 3D reconstruction.
We parameterize this policy as a deep network that maps input observations to actions and train using standard model-free reinforcement learning algorithms. Our primary contributions are in the construction of a sufficiently realistic simulation environment to support policy training, including careful reward shaping. We develop a synthetic tree generation pipeline based on formal grammars to procedurally generate realistic tree geometries \cite{lsystemsplants}. To avoid requiring photorealism or accurate depth simulation for thin structures, we adopt previous work that uses optical flow images, rather than RGB or depth images, for policy perception \cite{OpticalFlowAlex2022IROS}. Figure \ref{fig:tree_and_of} shows a comparison between a real and synthetic tree and their corresponding optical flows.

We validate our performance in simulation by placing cut points at 3 different locations on 1000 uniformly-sampled branch orientations (3000 total) -- achieving a success rate of 30\%. For comparison, an oracle RRT-Connect planner with perfect tree geometry achieves a success rate of 60\%. Further, the policy achieves a success rate of 35\% on 20 real-world trials, where the RRT-Connect planner largely fails because the reconstructed point-cloud tree geometry misses small branches. Finally, we perform a task space analysis to identify policy failure regions based on branch orientations.

Our primary contributions are as follows:


\begin{compactitem}[\hspace{5pt}--]
    \item A synthetic tree generation system that accounts for seasonal tree pruning and tie-downs as performed in modern orchard tree structures.
    \item A learned 6DOF visuomotor control policy capable of correctly orienting a cutter to a target branch cut-point using wrist-mounted camera images. 
    \item Zero-shot transfer from simulation to the real world by the use of optical flow.
    \item Identification of policy failure and success modes with respect to the working volume of the robot and varying branch orientations.
\end{compactitem}

\section{Related work}
\vspace{-4pt}
\label{sec:related}

Prior work in robotic pruning has focused on grapevine pruning using a traditional multi-step process. This involves: (i) constructing a point cloud from multiple multi-view depth images, (ii) determining which canes should be cut or trimmed, and (iii) employing a motion planner --- either RRT-Connect~\cite{CorbettDaviesStraddleVines,trimbot,SilwalBumblebee} or reinforcement learning (RL) policies~\cite{YandunRLPruning}—to guide the cutter to the designated cut points.

To construct the point clouds, depth information is obtained from cameras that utilize structured or coded lighting or time-of-flight sensors. However, these sensors see a degradation in performance when used in outdoor environments due to strong sunlight or surface scattering~\cite{lidaroutdoorbad}. As a result, controlling lighting is crucial for generating accurate point clouds. For example, Corbett-Davies et al. \cite{CorbettDaviesStraddleVines} straddles the grapevine rows within a box-like structure, creating a controlled environment with custom lighting. In contrast, Silwal et al. \cite{SilwalBumblebee} eliminates the need for such physical enclosures, instead using an advanced lighting system to regulate image exposure and obtain consistent depth.

Motion planning to reach pruning points is generally achieved using sampling-based planners, such as RRT-Connect, which treat the point cloud as the obstacle map. Due to the large number of collision objects in orchard environments, running these planners can be computationally intensive. To address this, Silwal et al.~\cite{SilwalBumblebee} reduce planning time by relaxing the collision detection constraints, allowing collisions with non-rigid parts of the tree during motion planning. In contrast, Yandun et al.~\cite{YandunRLPruning} replace sampling-based planning altogether with reinforcement learning, developing a policy that directly maps robot proprioception and occupancy grid data to motor torques.

Transitioning from these methods used for grapevines to tree pruning presents several challenges. First, the geometry of trees is more difficult to capture than that of grapevines due to the range of scales. Tree trunks range from 10 to 30 cm in diameter, with tertiary branches as thin as 1cm.  Second, these tertiary branches ``fill'' the space between support branches and are oriented in all directions, resulting in a large number of potential collisions. 
Given these limitations, You et al.~\cite{AlexUFOPruning} proposed a novel approach for cherry tree pruning that bypassed point clouds altogether. Instead, they use a learning-based visual servoing policy to guide the end-effector to the pruning point. This policy minimizes the image-based distance between the pruning point and the cutter's mouth; however, it assumed the branch was perpendicular to the cutter and did not avoid collisions.

We build on this approach, extending it to support 6DOF orientation and avoid (visible) obstacles.

\section{Simulating the Reaching Task for Robotic Tree Pruning}
\vspace{-4pt}
\label{sec:meth:sim}

To support visuomotor policy learning, we construct a simulator for robotic tree pruning that mimics the orchard environment. Our focus here is on producing plausible tree geometries that match tree architectures commonly used in commercial orchards -- but not necessarily generating photorealistic reproductions as our optical flow-based policy is not directly sensitive to image texture.

\vspace{4pt}\noindent \textbf{Generating Synthetic Trees.} There are several common modern planar tree orchard structures --- e.g., V-trellis, Upright Fruiting Offshoot, and Tall spindle~\cite{DeannaPruningDecisions,MattUFO} --- that all share the same characteristic structure. There is a post and wire structure to which trunks and/or branches are tied to induce specific shapes. This shaping is performed through annual cycles of tree growth followed by pruning and branch tying. While there are several existing packages for ``growing'' natural trees programmatically, none allow for the simulation of this shaping process required to generate realistic orchards. We extend an open-source software package (L-Py~\cite{lsystemsplants}), modifying its process to include cyclic pruning and tie-down phases during tree growth.

We focus on V-Trellis architectures in this work, however, the modified LPy software package can generate trees in any planar architecture. V-trellis is characterized by a vertical trunk with alternate trunks tilted at 15 degrees to form a V-shape.
Support branches extending from the trunk are tied to wires that run perpendicular to the trunk.  These wires are spaced 18 inches apart vertically, and each wire supports exactly one branch. The trees are spaced 4 feet apart on each side of the V such that these support branches are around 2 feet long. Figure \ref{fig:ur5_pruning_field} illustrates this V-Trellis structure, while Figure \ref{fig:tree_and_of} (Top) compares a real orchard tree and our generated trees.

We use a simple pruning strategy of selecting two branches for each unassigned wire (one for each side), removing any side branches coming off of the trunk that were not tied down in the previous year. These branches are tied down roughly every 6 inches for the length of the branch. For the resulting branch curvature, we model the branch as a cantilever beam with a load at the free end for each tie operation -- setting appropriate loads at the free end to flex the branch to the tie-down point and then solving for the resulting curvature.

With this framework, we can generate an arbitrary number of trees with stochastic variation and import them into the simulation environment. Though we have not formally verified that the tree geometry is statistically consistent with real orchards, we have compared the resulting geometry to our extensive scans of orchard trees and confirmed our modeling choices with a horticultural expert.

\vspace{4pt}\noindent \textbf{Simulator.} We create a PyBullet~\cite{pybullet} environment with the robot (UR5), trees, trellis wires, and posts. Each element in the environment is assigned a unique texture. Additionally, a textured wall is placed behind the tree to simulate the cluttered background of an orchard. Though not directly observed by the model, these textures are required to produce realistic optical flow images.

\vspace{4pt}\noindent \textbf{Pruner.} We add a custom cutter to the UR5 that has a camera in an ``eye-in-hand'' configuration (see Figure \ref{fig:ur5_pruning_field} (Right)). The robot is placed at a height consistent with a mobile platform and is oriented roughly orthogonal to the tree.


\vspace{4pt}\noindent \textbf{Defining Success.}  For a given pruning cutpoint on a branch, we consider an end-effector pose to be successful if the cutters have \emph{reached} the pruning point and are \emph{pointing} towards the branch with jaws \emph{perpendicular} to it, as illustrated in Figure~\ref{fig:success}. Specifically, the jaws of the cutter must be within 5cm of the cutpoint with the branch within the cutter mouth. The pointing and perpendicularity must be within 30 degrees of the nearest correctly aligned solution. These criteria meet the requirements of an existing admittance controller that can guide the cutter once it makes contact with the branch~\cite{PruningAdmitanceController2022ICRA}. Achieving an end-effector pose that satisfies the criteria requires avoiding branches to move sufficiently near the cutpoint and changing the roll, pitch, and yaw of the end-effector to orient for the cut correctly.

\section{Learning a Visuomotor Pruning Policy}
\vspace{-4pt}

We model vision-based control of a robotic pruner as a Partially-Observable Markov Decision Process (POMDP) defined by a state space $\mathcal{S}$, a continuous action space $\mathcal{A}$, environment transition dynamics $T: \mathcal{S}\times\mathcal{A}\rightarrow\mathcal{S}$, reward function $R: \mathcal{S}\rightarrow \mathbb{R}$, and observation function $\mathcal{O}: \mathcal{S} \rightarrow \Omega$ that maps states to partial observations $o \in \Omega$. Following standard practice in model-free reinforcement learning, we instantiate a parameterized policy $\pi_\theta: \mathcal{S} \rightarrow \mathcal{A}$ as a deep neural network that is optimized to maximize its expected discounted cumulative reward via PPO \cite{Schulman2017ProximalPO,stable-baselines3}.

\subsection{Observation and Action Spaces}
\label{sec:meth:state_space}
While the simulator provides full information about tree geometry and pose, the policy observation space is limited to 1) a goal that specifies the cut point, 2) optical flow imagery, and 3) robot proprioception.

\vspace{4pt}\noindent {\bf Cut-point Specification.} We include cut-point information in two forms -- i) an $(x, y, z)$ location of the cut point relative to the base of the robot and ii)  a 1-channel image ($224\times224$) rendering the 3D point as a 5 pixel-wide white circle in the current camera frame. 

\looseness=-1

\vspace{4pt}\noindent {\bf Optical Flow Image.} Rather than providing the policy with rendered RGB or depth images, we instead use an off-the-shelf optical flow model RAFT~\cite{raft_of} to calculate a 2-channel $224\times224$ optical flow image. This image provides the change in 2D image coordinates per pixel between rendered RGB images from the current and previous time steps. Prior work has shown that this approach offers strong sim-to-real generalization~\cite{OpticalFlowAlex2022IROS}. In real-world orchards, we have verified that optical flow captures the geometry of small branches better than a depth camera.
\looseness=-1

\vspace{4pt}\noindent {\bf Proprioception.} We include the 6-DOF pose, the end effector’s current velocity, and the robot’s joint angles in the state space. The end-effector’s pose is defined by its location
as $(x, y, z)$ coordinates and its orientation by 6D parameterization~\cite{6drotation}. The joint angles are represented using their sine and cosine values, providing the policy with information on the arm's proximity to singularities and self-collision. This results in a 27-dimensional proprioception state.
\looseness=-1

\vspace{4pt}\noindent {\bf Actions.} The policy outputs linear and angular velocities for the end-effector. Joint velocities are then computed by using the inverse Jacobian of the robot and applied for 0.5 seconds -- corresponding to a control frequency of 2 Hz. 
\looseness=-1

\subsection{Policy Architecture}
\label{sec:meth:policy}
We consider a fairly standard LSTM-based actor-critic architecture except for the addition of a privileged critic and an autoencoder to improve visual feature learning.

\vspace{4pt}\noindent \textbf{Visual encoder.} We stack the cutpoint and optical flow images into a single 3-channel $224\times224$ image $I_t$ and encode it with an 8-layer CNN with ReLU activations and a global average pool to produce a 72-dimensional latent state $z_t$. To provide additional supervision to the visual encoder \cite{Seo2022MaskedWM}, we also introduce a decoder CNN to reproduce the input image in an autoencoding fashion as $\hat{I}_t$. This is supervised with an L2 loss $||I_t - \hat{I}_t||_2$ in addition to the standard PPO objectives.

\vspace{4pt}\noindent \textbf{Actor.} The actor is a 1-layer LSTM followed by an MLP that at each time step takes an image encoding $z_t$ along with scalar features for proprioception, $(x, y, z)$ coordinates of the cutpoint in the world frame, and $(x, y, z)$ coordinates of the cutpoint in the end-effector frame. The resulting LSTM hidden state, which is of size 256, passes through a ReLU-activated MLP with progressively decreasing layer sizes from $512 \rightarrow 256 \rightarrow 128 \rightarrow 6$ which yields the predicted 6DOF velocity actions. Actions are parameterized as $tanh$-transformed Gaussians with means predicted by the policy and a shared variance parameter that is learned \cite{SACTanh}. These values are scaled to be within $\pm 0.1$m/s

\vspace{4pt}\noindent \textbf{Privileged Critic.} The critic shares the same image encoder as the actor and has an identical architecture, except: (i) it outputs a single scalar and (ii) it receives two additional scalar features corresponding to the pointing cosine similarity and perpendicular cosine similarity between the end-effector and the to-be-pruned branch. These inputs rely on privileged knowledge of the tree geometry and pose but are not available to the policy and are only used during training \cite{PrivilegedCriticPinto}. \looseness=-1
\subsection{Reward}
\label{sec:meth:reward}
\begin{figure}
    \centering    
    \vspace{2mm}
    \includegraphics[width=0.42\textwidth]{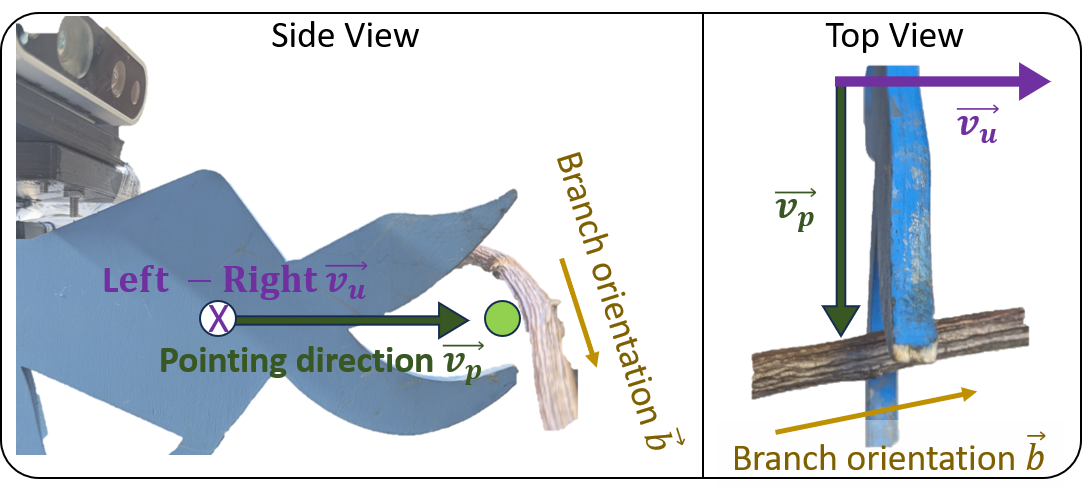}
    \vspace{-11pt}
    \caption{ Left: Side view of the cutter with the pointing direction $\vec{v_p}$ and left-right direction $\vec{v_u}$ labeled. The goal is to place the center of the cutter jaws (green circle) at the branch cut location, ensuring that the cutter is pointed at the branch with the jaws aligned. Right: Looking down, the jaws should be aligned with the branch ($\vec{v_u} \parallel \vec{b}$) and the cutter should be pointing to the branch ($\vec{v_p} \perp \vec{b}$). }
    \label{fig:success}
    \vspace{-11pt}
\end{figure}

Our reward function encourages \emph{reaching}, \emph{pointing} at, and being \emph{perpendicular} to the to-be-pruned branch. Further, we include a success reward when the branch enters the jaws of the cutter, collision penalties, and a slack reward to encourage efficient motions. When defining these rewards below, all points and vectors are given in the coordinate system of the robot base. Please refer to Figure~\ref{fig:success} as a visual guide.
\looseness=-1


\vspace{4pt}\noindent \textbf{Reaching Reward.} Let $p_e(t)$ be the point at the center of the cutter jaws at time $t$ and $p_g$ be the cut point. Then the \emph{reaching} reward at time $t$ is the change in distance between the cutpoint and end-effector at time $t$ and $t-1$:
\begin{equation}
R_{\mathtt{reach}}(t)  = \left\| p_{\mathrm{e}}(t{-}1) - p_{\mathrm{g}} \right\|_2 - \left\| p_{\mathrm{e}}(t) - p_{\mathrm{g}} \right\|_2
\end{equation}

\vspace{4pt}\noindent \textbf{Pointing Reward.} For orientation, we use cosine similarity ($C(v_1, v_2) \rightarrow [-1,1]$) to compare vectors. Let $R(t)$ be the rotation matrix that takes the unit vectors $\vec{x}, \vec{y}, \vec{z}$ to the left-right, up-down, and pointing vectors at time $t$. Let $\vec{b}$ be the branch orientation, $p_e(t)$ be the point at the center of the cutter jaws, and $p_g$ be the cut point. If the pointing vector $\vec{v_p}(t) = R(t)\vec{z}$ points at the branch, then the perpendicular projection of $p_g - p_e(t)$ on $\vec{b}$ denoted here as $\perpproj{(p_g - p_e(t))}{\vec{b}}$ should be in the same direction as $\vec{v_p(t)}$. As with the reaching reward, we provide the change in this cosine similarity as the pointing reward:
\begin{equation}
\begin{split}
    R_{\mathtt{point}}(t) = &\ C\left(\perpproj{({p}_{\mathrm{g}} - {p}_{\mathrm{e}}(t))}{\vec{b}}, \vec{v}_{\mathrm{p}}(t)\right) \\
    &- C\left(\perpproj{({p}_{\mathrm{g}} - {p}_{\mathrm{e}}(t{-}1))}{\vec{b}}, \vec{v}_{\mathrm{p}}(t{-}1)\right)
\end{split}
\end{equation}
\vspace{4pt}\noindent \textbf{Perpendicularity Reward.} The left-right vector $\vec{v_u}(t) = R(t) \vec{y}$ also needs to be aligned to the branch. As with prior rewards, we use the change in cosine similarity between $\vec{v_u}(t)$ and $\vec{b}$ as our perpendicularity reward:
\begin{equation}
R_{\mathtt{perp}}(t)  = \left|C(\vec{v_\mathrm{u}}(t), \vec{b})\right| - \left|C(\vec{v_\mathrm{u}}(t-1), \vec{b})\right|
\end{equation}
\vspace{4pt}\noindent \textbf{Collision Penalties.} If the robot collides with the environment, there is a negative reward $R_{\mathtt{col}}$ of -0.01 for a small branch and -0.1 for other rigid structures like trunks, primary branches, or poles.

\vspace{4pt}\noindent \textbf{Termination and Overall Reward.} We terminate an episode if 100 steps are reached or if the end-effector meets the success criteria for the given cutpoint. If successful, a terminal reward $R_{\mathtt{term}}$ of 2 is provided and 0 otherwise. The total reward at timestep $t$ is then the weighted sum of these individual rewards plus a constant slack $R_{\mathtt{slack}} = -0.1$ reward to encourage efficiency,
\begin{equation}
    \begin{split}
        R(t) = \alpha_{m} R_{\mathtt{reach}} + \alpha_{p1} R_{\mathtt{perp}} + \alpha_{p2} R_{\mathtt{point}}\\ + R_{\mathtt{term}} + R_{\mathtt{slack}} + R_{\mathtt{col}}    
    \end{split}
\end{equation}
\noindent where $\alpha_m = 5$, $\alpha_{p1} = 6$, $\alpha_{p2} = 2$ are empirically determined weighting coefficients.


\subsection{Generating Training Episodes}
\label{sec:meth:train}

An episode consists of a tree, a cutpoint, and a robot starting configuration. When generating training episodes, we would like high diversity in the location of cutpoints and the orientation of the corresponding to-be-pruned branch to ensure the policy is exposed to sufficient conditions to cover those it might encounter in deployment. 

To do so, we define a \emph{likely}-reachable region for the UR5 robot which includes all points 70cm to 95cm from the robot base, not including those below the horizontal plane or behind the vertical plane of the robot's base. This region is defined heuristically based on the robot's maximum reach with the cutters (95cm), the maximum outward-facing branch length (40cm), and the visibility of the tree for obstacle avoidance (closer than 30cm yielded poor views). This region would allow the robot (UR5) to reach pruning points on up to 3 trellis wires when placed in front of the tree in an orchard.\looseness=-1

For each training episode, we sample a cutpoint position uniformly within the \emph{likely}-reachable region as well as a random 3D orientation \cite{graphic_gems_3}. Given a bank of 100 generated trees, we search for prunable branches with that orientation ($\pm 10$ degrees) and then translate the corresponding branch's tree so that the branch is in the sampled position. This translation is constrained by the tree size so that the tree's appearance is not unnatural (e.g. tree floating mid air or the camera looking at empty spaces).  
Noise is added to the robot orientation ($\pm 5$ degrees over its yaw, pitch, and roll axis) and camera placement ($\pm 2$ degrees pan and tilt) to mimic real-world conditions when setting up a robot in the field.
\looseness=-1

\section{Experimental Results}
\vspace{-4pt}
\label{sec:experiments}

We evaluate our learned control policy both in simulation and on a proxy tree in the lab and compare it with a baseline RRT-Connect planner. 
The policy was trained on an NVIDIA 4090 Ti system for 8M timesteps and achieved a success rate of 24.21\% on training episodes. The policy in the real world was run real time using an NVIDIA 2080 Ti system.

\begin{figure}[t]
    \centering
    \includegraphics[width=0.49\textwidth, height=3.5cm]{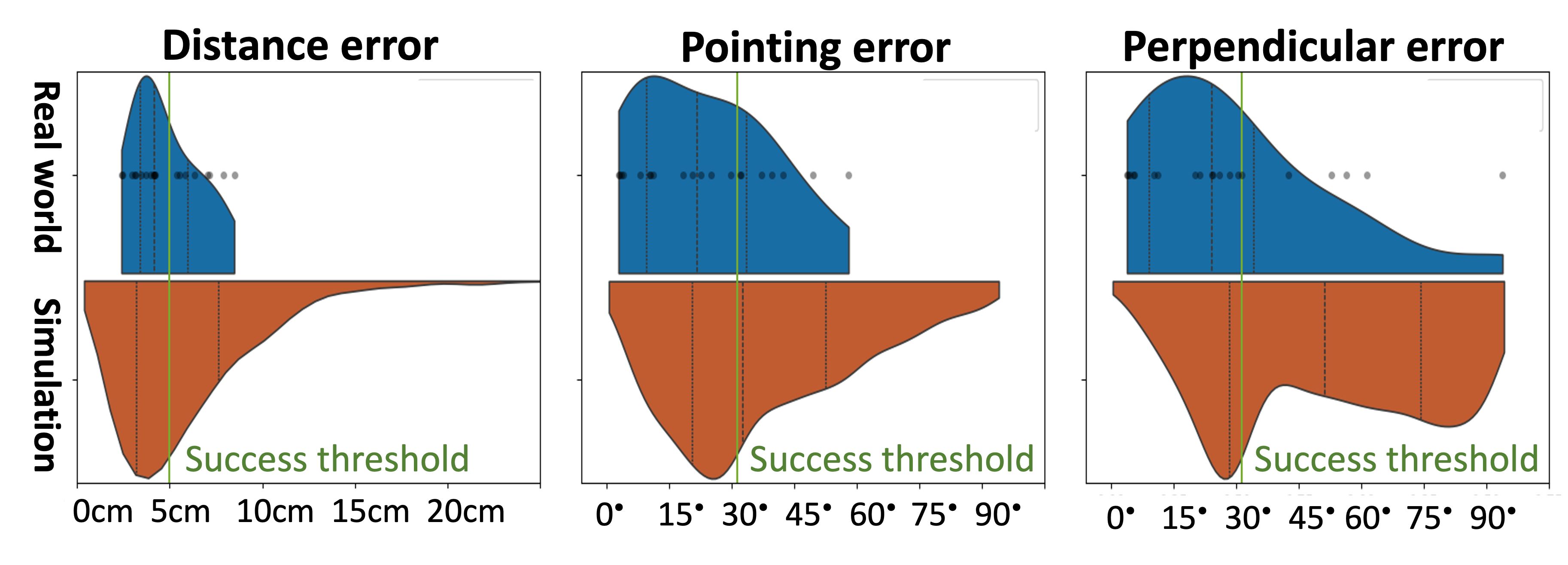}
    \vspace{-22pt}
    \caption{Error distributions for the (left) distance, (middle) pointing, and (right) perpendicular goals. Top: Real-world trials, with actual samples shown as dots (20). Bottom: 3,000 simulation trials.}
    \label{fig:error_violins}
    \vspace{-11pt}
\end{figure}

\subsection{Simulation Experiments}

For simulation experiments, we generate 3000 evaluation episodes from 10 novel trees. To provide coverage, the cutpoints are generated by uniformly sampling 1000 orientations~\cite{fibonaccisphere}, finding the corresponding branches, and translating them to 3 different locations within the likely-reachable space for each orientation. On these cutpoints, we run both the trained RL policy and an Oracle RRT-Connect planner:
\looseness=-1

\emph{RL-Policy}:
We executed the learned RL policy by performing the most-likely action predicted by the policy. For each cutpoint, the policy was run for 100 simulation steps or until the termination condition was reached.
\looseness=-1

\emph{Oracle RRT-Connect Planner}: The RRT-Connect planner is an oracle here because it uses the accurate tree mesh. It is important to note that obtaining such precise 3D meshes in the real world is extremely difficult, making this setup unrealistic for practical applications. Instead, this serves as the \emph{best-case performance} of a classical planner with perfect perception and tree-reconstruction quality. To generate the goal configuration for RRT-Connect, we sample 100 goal configurations for each cutpoint within the success region and consider it a success if a path exists to any of them.

\vspace{4pt}\noindent \textbf{Simulation Results.}
Across the 3000 simulation trials, the learned policy achieved success in 30.06\% of trials. Further, around 10\% of total trials included an unacceptable collision (e.g., with a support post). As an approximate upper bound, the oracle planning-based baseline with perfect perception achieved a collision-free 60.4\% success rate -- suggesting there remains significant headroom in this task. Orange density plots in Figure~\ref{fig:error_violins} show the distribution of reaching error (left), pointing error (middle), and perpendicularity error (right) for the trained policy. Vertical green lines indicate success thresholds. We find that pointing and perpendicularity pose more significant challenges to the policy than reaching. \looseness=-1
\begin{figure}[t]
    \centering
    \vspace{2.5mm}
    \includegraphics[width = 0.4\textwidth]{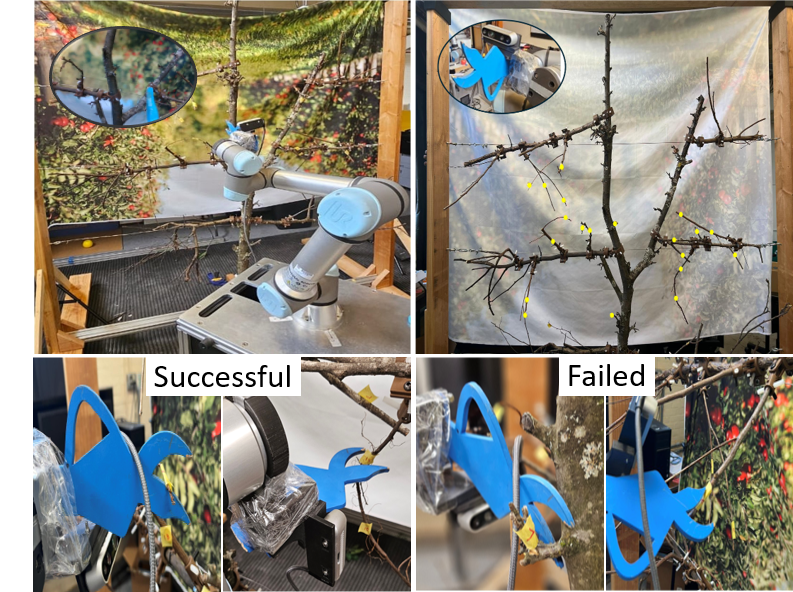}

    \caption{ Top-Left: The robot and physical tree in their starting configurations (backdrop is an image of an orchard). Inset is the image seen through the camera. Top-Right: The physical tree with target points for RL policy marked in yellow. The tree branches are real, but we use custom plastic connectors to control branch location and orientation. Inset is a picture of the cutter with an Intel RealSense D435 camera mounted on the wrist. Bottom-Left:  Examples of end-effector orientations for successful trials;  blue is the cutter. Bottom-Right: Examples of end-effector orientations for failed trials.}
    \label{fig:experimental_setup}
    \vspace{-11pt}
\end{figure}

\subsection{Real-World Experiments}
\label{sec:real_world_experiments}

\begin{figure}[t]
    \centering
    \vspace{2.5mm}
    \includegraphics[width=0.35\textwidth]{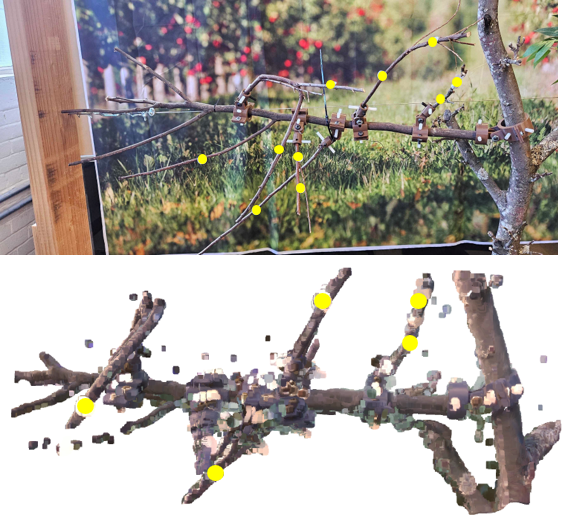}
    \vspace{-11pt}
    \caption{ Top : RBG image of the branch that was reconstructed with chosen cutpoints marked in yellow. Bottom: Corresponding point cloud}
    \label{fig:point_cloud}
    \vspace{-11pt}
\end{figure}

We conduct real-world trials on a proxy tree built in our laboratory. The tree is built from real apple tree branches but has custom connectors to attach branches at known angles. The tree is attached to trellis wires spaced the same as a real orchard as shown in Figure \ref{fig:experimental_setup} (Top-Right). We place the UR5 pointed at the tree (See Figure \ref{fig:experimental_setup} (Top-Left). We select 20 points to use as cutpoints (yellow marks in Figure \ref{fig:experimental_setup} and Figure \ref{fig:point_cloud}), manually driving the end-effector to those points to get the $(x, y, z)$ location in the robot's coordinate frame. For comparison with RRT-Connect, we build a point cloud of the tree using off-the-shelf 3D - reconstruction methods.

\looseness=-1

\emph{RL-Policy}:
We ran the robot 4x slower than in simulation (for safety) for a maximum of 3 minutes and 20 seconds or until the goal branch was within the cutters. 

\emph{RRT-Connect planner}: To construct the 3D point cloud we capture 7 point clouds with an Intel Realsense D435 camera and register them together using an off-the-shelf multi-scale colored ICP algorithm~\cite{ZhouOpen3d}. Due to the challenges in generating accurate point clouds, we only construct point cloud of one branch (Bottom left) from the RL trials in ideal lab conditions. Additional cut points were added for diversity in orientation for a total of 10. We executed the RRT-Connect planner only if the cutpoint was visible in the point cloud and let the planner run for a maximum of 60 seconds. The stitched point cloud and the corresponding RGB images can be seen in Figure \ref{fig:point_cloud}. The required poses were not calculated dynamically; instead, we manually supplied collision-free poses 5cm away that satisfied the success criteria to the planner.
\looseness=-1

\vspace{4pt}\noindent \textbf{Real-World Results.} For 20 real-world trials conducted with the RL-Policy, a success rate of 35\% was achieved matching the results in simulation -- suggesting high sim-to-real transferability. The blue density plots in Figure~\ref{fig:error_violins} show errors for the real-world trials. As these densities represent only 20 points each, we also plot the points as semi-transparent markers. We find similar trends as in simulation experiments, with larger orientation errors than distance errors. Figure \ref{fig:experimental_setup} (Bottom-Right and Bottom-Left) shows some successful and unsuccessful end-effector orientations at the end of the episodes. For both simulation and real trials for the RL policy, we observed that errors were due to cutpoints a) that were behind the tree and pointing away, b) that were directly pointing forward from the tree towards the UR5 arm, and c) that were flush with other branches. \looseness=-1

For the 10 trials conducted with RRT-Connect, 5 of the pruning points on tertiary branches were not captured in the point cloud, highlighting the challenges faced in point cloud reconstruction of thin tertiary branches. Of the remaining points, 2 trials were successful, and 3 led to non-convergence of the planner within the planning time. Our visuomotor policy outperforms point cloud based planners, this performance gap is due to erroneous point clouds that fail to capture smaller tertiary branches and wires and high clutter causing the RRT-Connect planner to fail.

\begin{figure}[t]
\centering
\includegraphics[width=0.42\textwidth]{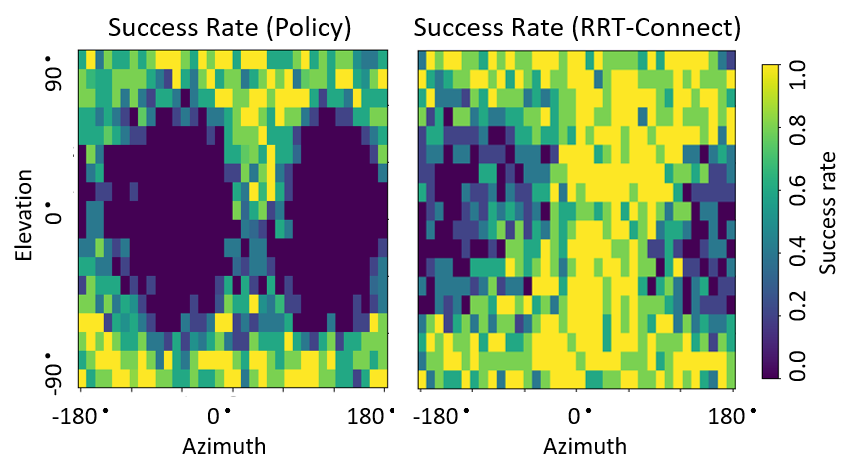}
\includegraphics[width=0.42\textwidth]{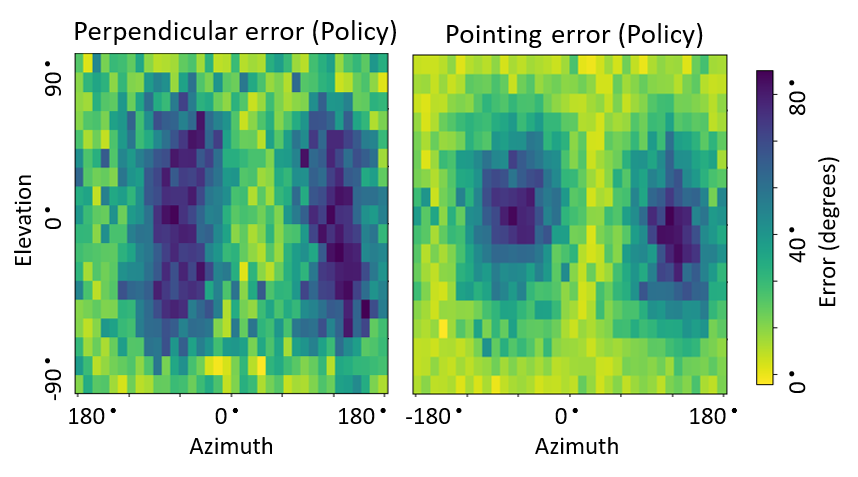}
\vspace{-11pt}
\caption{Failure analysis based on branch orientation. The Azimuth is the branch's orientation with respect to the ground plane; $\pm$ 90 degree is pointing toward/away from the robot. The Elevation axis is the vertical pointing direction, from pointing at the ground (-90 degree) to pointing at the sky (90 degree). Top: Success rates for policy and oracle RRT-Connect. Bottom: Pointing and perpendicular errors for RL policy.}
\label{fig:grid_analyis}
\vspace{-11pt}
\end{figure}

\section{Analysis}
\vspace{-4pt}
To study the error modes of the visuomotor policy more closely, we generate evaluation episodes in simulation with uniform coverage over to-be-pruned branch orientations. We represent branch orientation in terms of their azimuth and elevation and bin them into 10-degree cells over both axes for a total of $18\times36$ bins. For each cell, we randomly choose 5 branches with the corresponding orientation and place them in different 3D locations in the task space by translating their trees. To report the \emph{upper bound} success rate for the cutpoints, we run RRT-Connect planner with ground truth meshes. We report the mean success rate for each cell for both the RL policy and RRT-Connect. These results are shown in Figure~\ref{fig:grid_analyis} where elevation is the angle in the vertical plane (90 degree${\rightarrow}$straight up), and azimuth is the rotation in the horizontal plane (90 degree${\rightarrow}$straight at the robot).

Looking at success rates (Figure \ref{fig:grid_analyis} (Top)), we find highly-structured error patterns for the oracle-based RRT-Connect and similar but pronounced patterns for our policy. For RRT-Connect, the two dark regions of failures are centered at 0 degree elevation and $\pm$ 145 degree azimuth -- suggesting that it is impossible to get to these poses. This can be attributed to the asymmetric structure of the UR5 arm and self-collisions due to the addition of the pruner on one side. For our RL-policy (Figure \ref{fig:grid_analyis} (Bottom)), the primary cause of orientation failures is the inability to position the end-effector to point toward these branches. This issue could be caused by exploration challenges in this region of large negative rewards due to self-collisions and high joint velocities around positions due to singularities in the robot Jacobian.


\section{Conclusion}
\vspace{-4pt}

We presented a pipeline to train a closed-loop, optical-flow based controller to handle dormant pruning in modern planar orchards. We demonstrate that this policy learned in simulation using our novel synthetic tree generation system can transfer to a real robotic platform and attain similar performance as in simulation. This task is made challenging by visual and physical clutter from branches and a wide range of end-effector goal orientations. Despite these challenges, the policy achieves 50\% performance of an oracle planner.

\section*{ACKNOWLEDGMENT}
This work was supported by the AI Institute for Agricultural AI for Transforming Workforce and Decision Support (AgAID), funded through the National Science Foundation (NSF) and the U.S. Department of Agriculture's National Institute of Food and Agriculture (USDA-NIFA) under the AI Research Institutes program, award No. 2021-67021-35344.
\looseness=-1

\bibliographystyle{ieeetr}
\bibliography{bibliography}
\end{document}